\documentclass[pdflatex,sn-mathphys-num,iicol]{sn-jnl}

\usepackage{graphicx}%
\usepackage{multirow}%
\usepackage{amsmath,amssymb,amsfonts}%
\usepackage{amsthm}%
\usepackage{mathrsfs}%
\usepackage[title]{appendix}%
\usepackage{xcolor}%
\newcommand{\chg}[1]{\textcolor{gray}{\textit{(#1)}}}
\usepackage{textcomp}%
\usepackage{makecell}
\usepackage{manyfoot}%
\usepackage{booktabs}%
\usepackage{algorithm}%
\usepackage{algorithmicx}%
\usepackage{algpseudocode}%
\usepackage{listings}%

\theoremstyle{thmstyleone}%
\theoremstyle{thmstyletwo}%

\theoremstyle{thmstylethree}%

\begin{document}

\title[Rethinking Underwater Image Enhancement]{Rethinking Underwater Image Enhancement: Learning through Text-Guided Degradation Curriculum}

\author[1]{\fnm{Hongri} \sur{Tian}}
\equalcont{Hongri Tian and Wenxue Li are co-first authors.}
\author[1]{\fnm{Wenxue} \sur{Li}}
\equalcont{Hongri Tian and Wenxue Li are co-first authors.}
\author[1]{\fnm{Tian} \sur{Ye}}
\author[1]{\fnm{Sixiang} \sur{Chen}}
\author[1]{\fnm{Zongxian} \sur{Su}}
\author[1]{\fnm{Lujia} \sur{Wang}}
\author[2]{\fnm{Xianying} \sur{Wang}}
\author[1]{\fnm{Xinhu} \sur{Zheng}}
\author[3]{\fnm{Gang} \sur{Hua}}
\author*[1,4]{\fnm{Lei} \sur{Zhu}}\email{leizhu@hkust-gz.edu.cn}

\affil[1]{\orgname{The Hong Kong University of Science and Technology (Guangzhou)}, \orgaddress{\city{Guangzhou}, \country{China}}}

\affil[2]{\orgname{China Geological Survey}, \orgaddress{\city{Beijing}, \country{China}}}

\affil[3]{\orgname{Director of Applied Science,  Amazon.com, Inc.}, \orgaddress{\country{U.S.}}}

\affil[4]{\orgname{The Hong Kong University of Science and Technology}, \orgaddress{\city{Hong Kong SAR}, \country{China}}}

\abstract{Existing paired clean–degraded datasets have advanced learning-based underwater image enhancement, but they remain either limited in scale or difficult to acquire, and still fail to adequately cover the diverse degradations encountered in real-world underwater environments. Such scarce and insufficient data ultimately restrict the generalization of underwater image enhancement models, highlighting the need for realistic and controllable degradation modeling to better capture real underwater conditions and improve enhancement robustness.
To this end, we propose a unified degradation-aware training framework for the underwater image enhancement task. A Text-Conditioned Degradation Augmentation Pipeline uses a text-guided diffusion model to synthesize diverse and semantically consistent degraded samples, providing paired data to support the subsequent curriculum learning strategy. The proposed Progressive Degradation Curriculum organizes training across multiple attributes and difficulty levels to promote balanced learning, and a Transmission-Guided One-Step Diffusion Model leverages this progressive process to better address the wide range of degradation severity and restoration ambiguity in underwater images.
Extensive experiments on multiple underwater benchmarks demonstrate that our proposed method achieves superior performance. Our code and models will be released upon the publication of this work.}


\maketitle

\begin{figure}
\centering{\includegraphics[width=0.47\textwidth]{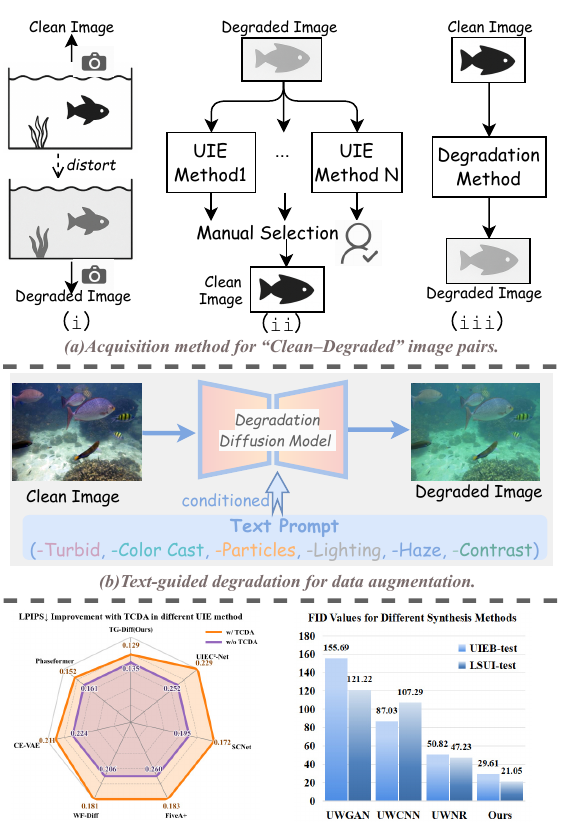}}
\caption{\textit{Top}: Illustration of three common routes for constructing paired underwater datasets. \textit{Medium}: Proposed text-driven degradation diffusion paradigm for controllable underwater degradation synthesis. \textit{Bottom}: Quantitative results of our method on performance improvement and synthesis quality.}
\label{Fig.teaser}
\end{figure}

\section{Introduction}
Underwater Image Enhancement (UIE) aims to recover clear, color-consistent, and detail-preserving images from degraded underwater observations.
Unlike uniform degradations in common image restoration tasks, underwater degradations are highly heterogeneous and scene-dependent, where they vary with water type, depth, illumination, and particle density, and often exhibit complex spatial coupling among these factors. Consequently, the underlying degradation distribution is extremely difficult to model or approximate with a fixed prior, making UIE an intrinsically ill-posed and data-sensitive problem.

Over the past decade, a wide range of enhancement approaches have been explored, spanning from physics-based restoration guided by analytical priors~\cite{drews2013transmission,galdran2015automatic,chiang2011underwater} to deep learning-based methods trained on paired datasets~\cite{li2020underwater,li2021underwater,wang2021uiec,zhang2022underwater,wei2022uhd,li2019underwater}. 
While the former provides physical interpretability, it fails to generalize across diverse water types and illumination conditions. 
The latter achieves stronger visual quality through elaborately designed networks but relies heavily on limited paired data. 

Acquiring paired underwater data is inherently challenging because capturing the same scene under both degraded and clean conditions is almost impossible in dynamic environments. As illustrated in Figure~\ref{Fig.teaser}-Top, such pairs are commonly obtained through \textit{(i)} controlled water-tank capture~\cite{duarte2016dataset}, \textit{(ii)} pseudo-reference generation by manually selecting outputs from multiple UIE methods~\cite{li2019underwater,peng2023u}, or \textit{(iii)} synthetic rendering with simplified underwater formation models~\cite{islam2020fast,li2017watergan,ye2022underwater,li2020underwater}. The first two routes provide visually realistic supervision but are costly and difficult to scale, whereas the third enables scalable generation but often fails to reproduce the complex lighting and scattering effects of natural underwater scenes.

These limitations reveal that the main bottleneck of UIE lies not only in model architecture but also in the lack of realistic and controllable degradation modeling for generating diverse training variants.
Moreover, underwater degradations vary drastically in strength and composition—ranging from light color cast to dense scattering and mixed turbidity. 
This motivates us to rethink UIE — not as direct restoration, but as \textit{learning from degradation}, where controllable, text-guided degradation modeling provides structured, degradation-aware supervision for enhancement learning.
In addition, we argue that if degradations can be explicitly modeled and controlled, they can serve as a meaningful curriculum for enhancement, where the model gradually adapts to increasingly challenging degradations across multiple attributes and severity levels, which encourages more generalized enhancement performance.

To this end, we revisit underwater enhancement from a new perspective-leveraging the generative capability of the diffusion model to explicitly \textit{model} and \textit{control} underwater degradations (Figure~\ref{Fig.teaser}-Medium). 
Diffusion-based generative models~\cite{labs2025flux} provide an expressive latent space in which degradations can be synthesized and conditioned on semantic cues such as turbidity, illumination, or particle density. 
This text-conditioned formulation enables fine-grained control over degradation attributes and supports large-scale data augmentation aligned with real underwater characteristics.
The generated degradations further provide difficulty-aware cues that guide enhancement models to learn progressively under well-structured degradation distributions, yielding more stable optimization and visually coherent restoration across diverse underwater conditions.
Moreover, we develop a Transmission-Guided One-Step Diffusion enhancer that injects a transmission prior into the latent space and performs a single deterministic reverse update, providing transmission-guided restoration with lower sampling cost than multi-step diffusion methods.
Together, these designs establish a cohesive paradigm that unifies controllable degradation modeling and physically guided enhancement, offering a new perspective on the UIE task.

Overall, our contributions can be summarized as follows:
\begin{itemize}
    \item We introduce a Text-Conditioned Degradation Augmentation (TCDA) pipeline that leverages a text-guided diffusion model to synthesize controllable degraded variants of clean images, enabling fine-grained attribute/severity control for task-aligned data augmentation.
    \item {We employ a progressive degradation curriculum learning that gradually increases training difficulty, stabilizing optimization and improving the model’s robustness.}
    \item We design a Transmission-Guided One-step Diffusion model (TG-Diff) that embeds a transmission prior into the latent space, achieving high-quality restoration with greater inference efficiency than multi-step diffusion methods.
    \item Extensive experiments on multiple underwater benchmarks demonstrate that our method achieves state-of-the-art performance.
\end{itemize}

\begin{figure*}[t]
\centering{\includegraphics[width=1.0\textwidth]{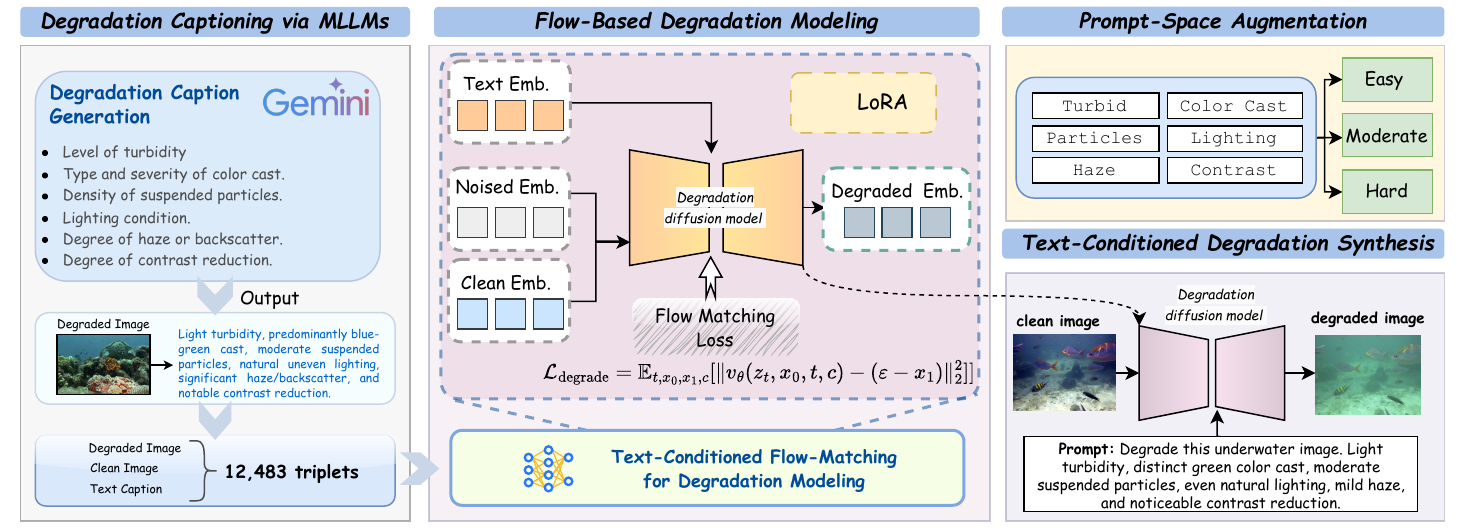}}
\caption{Overview of the proposed Text-Conditioned Degradation Augmentation (TCDA) pipeline.
It consists of four stages: (1) Degradation Captioning via MLLMs for generating fine-grained textual descriptions of underwater degradations, (2) Flow-Based Degradation Modeling for learning text-conditioned mappings between clean and degraded representations, (3) Prompt-Space Augmentation for enriching degradation prompts with varying difficulty levels, and (4) Text-Conditioned Degradation Synthesis for controllable degradation generation to support curriculum learning.}
\label{Fig.datapipe}
\end{figure*}

\section{Related Work}

\subsection{Underwater Image Enhancement}
Traditional UIE methods are typically built upon the underwater image formation model that characterizes light scattering and absorption.
Early works such as underwater dark channel prior~\cite{drews2013transmission}, red channel prior~\cite{galdran2015automatic}, and blurriness-based estimation~\cite{chiang2011underwater} estimate the transmission map and background light to remove haze and recover true colors. 
More recent prior-based methods further exploit illumination characteristics to handle non-uniform lighting and color degradation~\cite{hou2023non,zhou2025underwater}.
However, their performance heavily depends on accurate estimation of water parameters and scene depth, limiting their robustness across varying underwater conditions~\cite{liu2020real}.
Deep learning has revolutionized UIE by providing strong nonlinear modeling and robust feature representation capabilities.
Early convolutional methods~\cite{li2020underwater,li2021underwater,wang2021uiec,zhang2022underwater,wei2022uhd} employed encoder--decoder or multi-branch architectures to learn direct mappings from degraded to clean images, achieving improved contrast and color fidelity through illumination modeling and multi-level feature learning. Adversarial and generative frameworks~\cite{fabbri2018enhancing,wang2019uwgan} further enhanced perceptual realism but often introduced instability and color over-saturation. 
Moreover, some works use contrastive learning~\cite{zhou2024hclr},
domain priors, and external knowledge to improve UIE generalization
and controllability.
PhaseFormer~\cite{khan2025phaseformer} introduces phase-based attention to preserve structural cues, while CLIP-UIE~\cite{liu2025underwater} utilizes semantic guidance from vision-language models. Semantic-Guided Diffusion~\cite{zhou2025semantic} leverages multimodal cues from LLaVA and Zip-CLIP to generate semantic heatmaps for diffusion guidance.

Recent studies highlight the role of generative priors and degradation modeling in diffusion-based image restoration~\cite{li2025diffusion}.
Underwater-specific approaches, including WF-Diff~\cite{zhao2024wavelet} and SLURPP~\cite{wu2025single}, exploit diffusion models for underwater enhancement.
Related diffusion-based restoration methods further explore degradation-model optimization~\cite{fei2026generative}, global brightness alignment~\cite{guo2026p}, and blur-decoupled learning~\cite{cheng2026bd}.

Despite these advances, the effectiveness of related UIE methods remains largely constrained by the availability and diversity of training data.

\subsection{Underwater Image Synthesis}
The scarcity of paired and diverse underwater data has motivated the research on underwater image synthesis. 
Early physics-based methods simulate light attenuation using simplified water models but fail to reproduce complex scattering~\cite{ye2022underwater,li2020underwater}.
Alternatively, some datasets create pseudo-references by applying multiple UIE methods to a degraded image and having a user select the best outcome as the ``clean'' target~\cite{li2019underwater}. While visually plausible, this approach is fundamentally limited as it provides no insight into the forward degradation process itself. It merely offers a single, fixed enhancement target for each input, failing to model the diverse and controllable degradation attributes (e.g., water type, turbidity) that are crucial for robust generalization. 
GAN-based approaches~\cite{li2017watergan,islam2020fast} improve realism but suffer from domain bias and unstable training.
Recent diffusion-driven frameworks, including Atlantis++~\cite{zhang2026atlantis++} and TIDE~\cite{lin2025unified}, enable text- or depth-conditioned underwater scene synthesis.
However, these methods focus on generating underwater scenes rather than modeling controllable degradations. 
Our work uniquely fills this gap by introducing a controllable, text-driven degradation diffusion model that synthesizes diverse and realistic underwater degradations, providing richer supervision and significantly enhancing the generalization of UIE models.

\section{Method}
Our methodology addresses the challenges of underwater image enhancement through a two-stage framework. First, to overcome the bottleneck of limited diversity and fixed degradation patterns in existing datasets, we introduce a Text-Conditioned Degradation Augmentation (TCDA) pipeline. This pipeline synthesizes a vast and diverse corpus of degraded images with controllable degradation attributes specified by fine-grained text prompts. 
{Second, leveraging this augmented data, we employ a progressive degradation curriculum to train a Transmission-Guided One-Step Diffusion (TG-Diff) model, designed for transmission-guided image restoration with greater inference efficiency than multi-step diffusion methods.}

\subsection{Text-Conditioned Degradation Augmentation}
As illustrated in Figure~\ref{Fig.datapipe}, our TCDA pipeline establishes a controllable mapping between clean images, their degraded counterparts, and textual degradation prompts.

\subsubsection{Degradation Captioning via MLLMs}
We first collect a large-scale dataset of 12,483 degraded--clean image pairs from four representative underwater datasets: UIEB~\cite{li2019underwater}, EUVP~\cite{islam2020fast}, LSUI~\cite{peng2023u}, and 3D TURBID~\cite{duarte2016dataset}. To ensure fair evaluation, only the training subsets of UIEB and LSUI are used. 

To create rich, structured textual representations of the visual degradations, we employ a multimodal large language model (Gemini 2.5-Flash~\cite{comanici2025gemini}) to generate fine-grained captions. These captions describe six core degradation attributes: turbidity level, color-cast type and severity, density of suspended particles, lighting conditions, degree of haze or backscatter, and extent of contrast reduction. 

{To obtain more consistent captions, we guide the MLLM to describe degradation cues only, while excluding scene content.} 
This process yields a dataset of triplets $\{I^i_{\text{deg}}, I^i_{\text{clean}}, T^i_{\text{deg}}\}_{i=1}^N$, where the text $T^i_{\text{deg}}$ provides fine-grained supervision that directly links semantic attributes to visual degradation patterns, forming the foundation for our text-conditioned diffusion modeling.{ Details about the prompt design are provided in Supplementary Sec.~2.1.}

\begin{figure*}[t]
\centering{\includegraphics[width=1.0\textwidth]{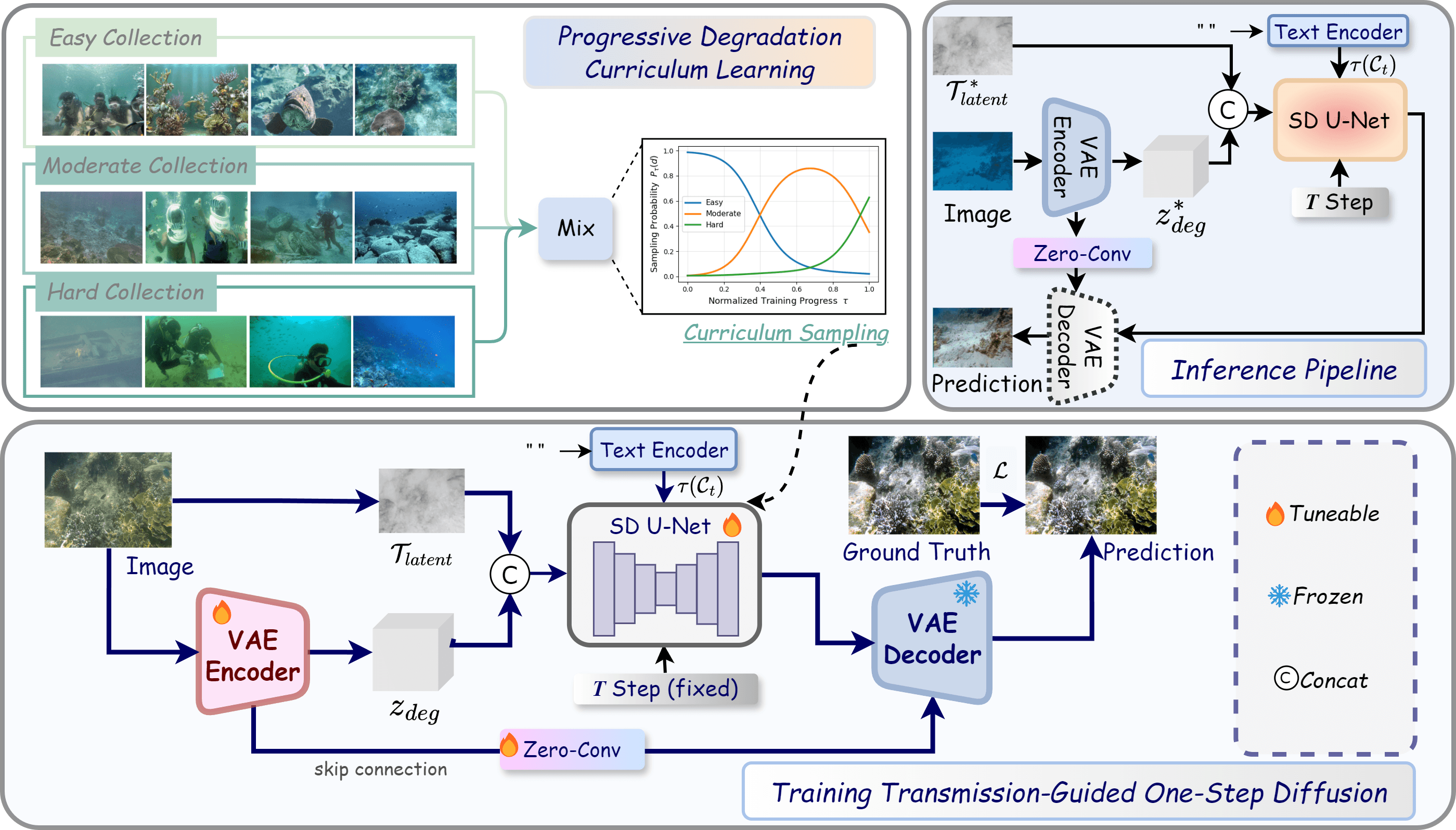}}
\caption{Overview of the proposed UIE framework. 
The model learns progressive degradation-aware enhancement through curriculum sampling and transmission-guided one-step diffusion, combining physical priors and text-conditioned cues for controllable restoration with greater inference efficiency than multi-step diffusion methods.
}
\label{Fig.method-frame}
\end{figure*}

\subsubsection{Flow-Based Degradation Modeling}
Building upon the constructed triplets, we develop a text-conditioned diffusion framework to learn the mapping from clean to degraded representations. To model this directed transformation, we adopt a flow-based diffusion framework based on FLUX.1 Kontext~\cite{labs2025flux}. 

Unlike traditional denoising-based models, flow-matching directly learns the vector field from a source to a target distribution, offering a more stable and efficient training objective for this task. Specifically, the clean image $I_{\text{clean}}$ and degraded image $I_{\text{deg}}$ are encoded into compact latent representations $x_0$ and $x_1$ via a pre-trained VAE encoder. Concurrently, the textual description $T_{\text{deg}}$ is encoded by frozen dual text encoders (T5-XXL~\cite{raffel2020exploring} and CLIP-L~\cite{radford2021learning}) to obtain an embedding $c$. The model learns a conditional velocity field $v_\theta(z_t, x_0, t, c)$ that drives the transformation of $x_0$ towards $x_1$ under text guidance. We adopt a flow-matching objective as
\begin{equation}
\mathcal{L}_{\text{degrade}} =
\mathbb{E}_{t,x_0,x_1,c}
\big[\|v_\theta(z_t, x_0, t, c) - (\varepsilon - x_1)\|_2^2\big],
\end{equation}
where $z_t$ denotes the intermediate latent state at timestep $t$, and $\varepsilon \sim \mathcal{N}(0,1)$ represents the stochastic noise.
This formulation enables the model to learn a text-conditioned degradation vector field that reflects both the deterministic transformations described by the text and the stochastic variations inherent to real-world degradations.
To enhance training efficiency and reduce computational overhead, we adopt parameter-efficient fine-tuning with LoRA~\cite{hu2021lora}.

\subsubsection{Controllable Degradation Synthesis}
To fully leverage our text-conditioned degradation model, we construct a structured prompt bank. {Specifically, instead of relying on unconstrained free-form prompts, we organize the prompt space around six degradation attributes, including turbidity, color cast, particles, lighting, haze/backscatter, and contrast. For each attribute, we define phrases with different severity levels and group them into three difficulty tiers, i.e., easy, moderate, and hard, so that the resulting prompts are both semantically coherent and explicitly controllable}. Each prompt is a tuple $T_{\text{deg}} = (a_1, \ldots, a_K, d)$, detailing key attributes $\{a_k\}$ and a difficulty level $d \in \{\texttt{easy}, \texttt{moderate}, \texttt{hard}\}$. During synthesis, given a clean image $I_{\text{clean}}$ and any such prompt $T_{\text{deg}}$, our model generates a corresponding degraded image $I_{\text{deg}}$. This text-driven paradigm not only enables fine-grained semantic control over degradation types and intensities but also provides the multi-difficulty data crucial for our progressive enhancement training, as detailed next. Representative prompt-conditioned degradation results are shown in Figure~\ref{Fig.teaser2}. {Detailed design is described in Supplementary Sec.~2.2.}

\begin{figure}
\centering{\includegraphics[width=0.50\textwidth]{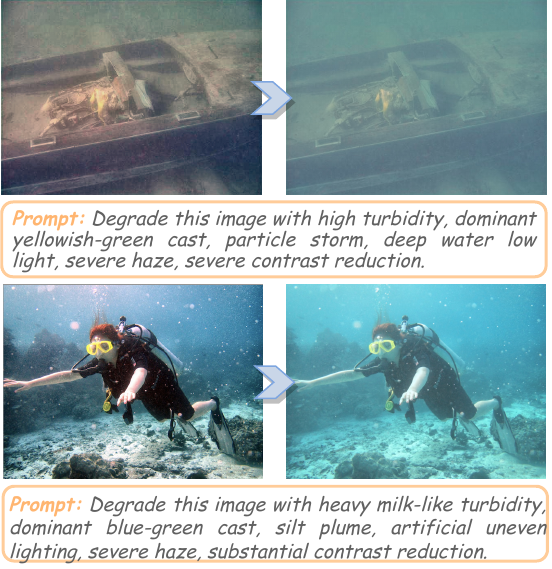}}
\caption{Examples of learned degradation generation results. Given a clean image and a textual prompt explicitly describing degradation attributes (e.g., turbidity, color cast, lighting, haze), our model synthesizes degraded counterparts that accurately reflect the specified visual conditions and severity levels.}
\label{Fig.teaser2}
\end{figure}

\renewcommand{\arraystretch}{1.2}
\begin{table*}[t]
\centering
\setlength{\tabcolsep}{1.5pt}
\caption{Quantitative comparison on UIEB, C60, and U45. Each method is evaluated with its original training data and with TCDA augmentation. Parentheses show the performance change. Best results are in bold.}
\label{compare_main}
\resizebox{\textwidth}{!}{
\begin{tabular}{l|ccccc|cc|cc}  
\toprule
\multirow{2}{*}{Methods} 
  & \multicolumn{5}{c|}{UIEB}
  & \multicolumn{2}{c|}{C60}
  & \multicolumn{2}{c}{U45} \\
\cmidrule(lr){2-6}\cmidrule(lr){7-8}\cmidrule(lr){9-10}
  & SSIM$\uparrow$ & PSNR$\uparrow$ & LPIPS$\downarrow$ & UCIQE$\uparrow$ & UIQM$\uparrow$
  & UCIQE$\uparrow$ & UIQM$\uparrow$
  & UCIQE$\uparrow$ & UIQM$\uparrow$ \\
\midrule
UIEC$^{2}$\text{-}Net~\cite{wang2021uiec}
& 0.880 & 17.961 & 0.252 & 0.526 & 3.766
& 0.487 & 2.149
& 0.535 & 4.231 \\
~ + TCDA
& 0.880 {\chg{+0.000}} & 19.020 {\chg{+1.059}} & 0.229 {\chg{-0.023}} & 0.550 {\chg{+0.024}} & 3.695 {\chg{-0.071}}
& 0.525 {\chg{+0.038}} & 2.146 {\chg{-0.003}}
& 0.554 {\chg{+0.019}} & 4.156 {\chg{-0.075}} \\
\midrule
SCNet~\cite{fu2022underwater}
& 0.907 & 19.560 & 0.195 & 0.585 & 3.869
& 0.566 & 2.498
& 0.570 & 4.155 \\
~ + TCDA
& 0.909 {\chg{+0.002}} & 20.998 {\chg{+1.438}} & 0.172 {\chg{-0.023}} & 0.593 {\chg{+0.008}} & 3.925 {\chg{+0.056}}
& 0.551 {\chg{-0.015}} & 2.452 {\chg{-0.046}}
& 0.578 {\chg{+0.008}} & 4.163 {\chg{+0.008}} \\
\midrule
FiveA+~\cite{jiang2023five}
& 0.903 & 18.671 & 0.260 & 0.558 & 3.811
& 0.531 & 2.229
& 0.546 & 4.186 \\
~ + TCDA
& 0.907 {\chg{+0.004}} & 20.516 {\chg{+1.845}} & 0.183 {\chg{-0.077}} & 0.586 {\chg{+0.028}} & 3.845 {\chg{+0.034}}
& 0.548 {\chg{+0.017}} & 2.340 {\chg{+0.111}}
& 0.563 {\chg{+0.017}} & 4.080 {\chg{-0.106}} \\
\midrule
WF\text{-}Diff~\cite{zhao2024wavelet}
& 0.901 & 19.984 & 0.206 & 0.590 & 4.075
& 0.551 & 2.404
& 0.568 & 4.140 \\
~ + TCDA
& 0.914 {\chg{+0.013}} & 20.622 {\chg{+0.638}} & 0.181 {\chg{-0.025}}
& 0.592 {\chg{+0.002}} & 3.902 {\chg{-0.173}}
& 0.553 {\chg{+0.002}} & 2.384 {\chg{-0.020}}
& 0.570 {\chg{+0.002}} & 4.167 {\chg{+0.027}} \\
\midrule

CE\text{-}VAE~\cite{pucci2025vae}
& 0.818 & 20.032 & 0.224 & 0.598 & 4.098
& 0.555 & 2.756
& 0.590 & 4.322 \\
~ + TCDA
& 0.857 {\chg{+0.039}} & 19.786 {\chg{-0.246}} & 0.211 {\chg{-0.013}} & 0.601 {\chg{+0.003}} & \textbf{4.110} {\chg{+0.012}}
& 0.559 {\chg{+0.004}} & \textbf{2.786} {\chg{+0.030}}
& 0.595 {\chg{+0.005}} & 4.280 {\chg{-0.042}} \\
\midrule
Phaseformer~\cite{khan2025phaseformer}
& 0.917 & 21.643 & 0.161 & 0.594 & 3.938
& 0.543 & 2.431
& 0.588 & 4.223 \\
~ + TCDA
& 0.923 {\chg{+0.006}} & 22.045 {\chg{+0.402}} & 0.152 {\chg{-0.009}} & 0.607 {\chg{+0.013}} & 3.946 {\chg{+0.008}}
& 0.553 {\chg{+0.010}} & 2.483 {\chg{+0.052}}
& 0.593 {\chg{+0.005}} & 4.219 {\chg{-0.004}} \\
\midrule
\textbf{TG\text{-}Diff (Ours)}
& 0.917 & 21.723 & 0.135 & 0.598 & 4.040
& 0.557 & 2.572
& 0.603 & 4.326 \\
~ + TCDA
& \textbf{0.927} {\chg{+0.010}} & \textbf{22.341} {\chg{+0.618}} & \textbf{0.129} {\chg{-0.006}} & \textbf{0.613} {\chg{+0.015}} & 3.982 {\chg{-0.058}}
& \textbf{0.572} {\chg{+0.015}} & 2.633 {\chg{+0.061}}
& \textbf{0.611} {\chg{+0.008}} & \textbf{4.339} {\chg{+0.013}} \\
\bottomrule
\end{tabular}}
\end{table*}

\subsection{Progressive Degradation Curriculum Learning}
Directly training a restoration model on the full, diverse spectrum of degradations generated by TCDA can lead to instability and slow convergence. To address this, we introduce a progressive degradation curriculum that leverages the difficulty levels defined in our prompt bank. This curriculum gradually increases the degradation complexity during training, allowing the model to first master fundamental color and tone restoration before tackling more complex scattering and contrast loss.

Specifically, for each difficulty level $d \in$ \{\texttt{easy}, \texttt{moderate}, \texttt{hard}\}, we formulate a sampling schedule that dynamically adjusts the selection probability $P_{\tau}(d)$ as a function of the normalized training progress $\tau \in [0,1]$:
\begin{equation}
    P_{\tau}(d) = 
    \frac{\exp\big(\alpha_p \cdot s_d(\tau)\big)}{\sum_{d'} \exp\big(\alpha_p \cdot s_{d'}(\tau)\big)},
\end{equation}
where $\alpha_p$ is a temperature factor controlling the transition sharpness, which we set to 5.0 in our experiments to ensure a smooth yet effective curriculum, and $s_d(\tau)$ is the preference score for difficulty $d$. To realize a smooth shift, we instantiate the scores as:
\begin{equation}
\begin{aligned}
    s_{\text{easy}}(\tau)   &= \cos^2\!\left(\frac{\pi \tau}{2}\right), \\
    s_{\text{moderate}}(\tau) &= \sin(\pi \tau), \\
    s_{\text{hard}}(\tau)   &= \sin^2\!\left(\frac{\pi \tau}{2}\right),
\end{aligned}
\end{equation}
This formulation enables a smooth and interpretable evolution of the training data distribution, allowing the model to gradually adapt to increasingly complex degradations. As shown in Figure~\ref{Fig.method-frame}, the continuous evolution of sampling probabilities promotes progressive robustness across diverse underwater conditions.

\begin{figure*}[t]
\centering{\includegraphics[width=0.98\textwidth]{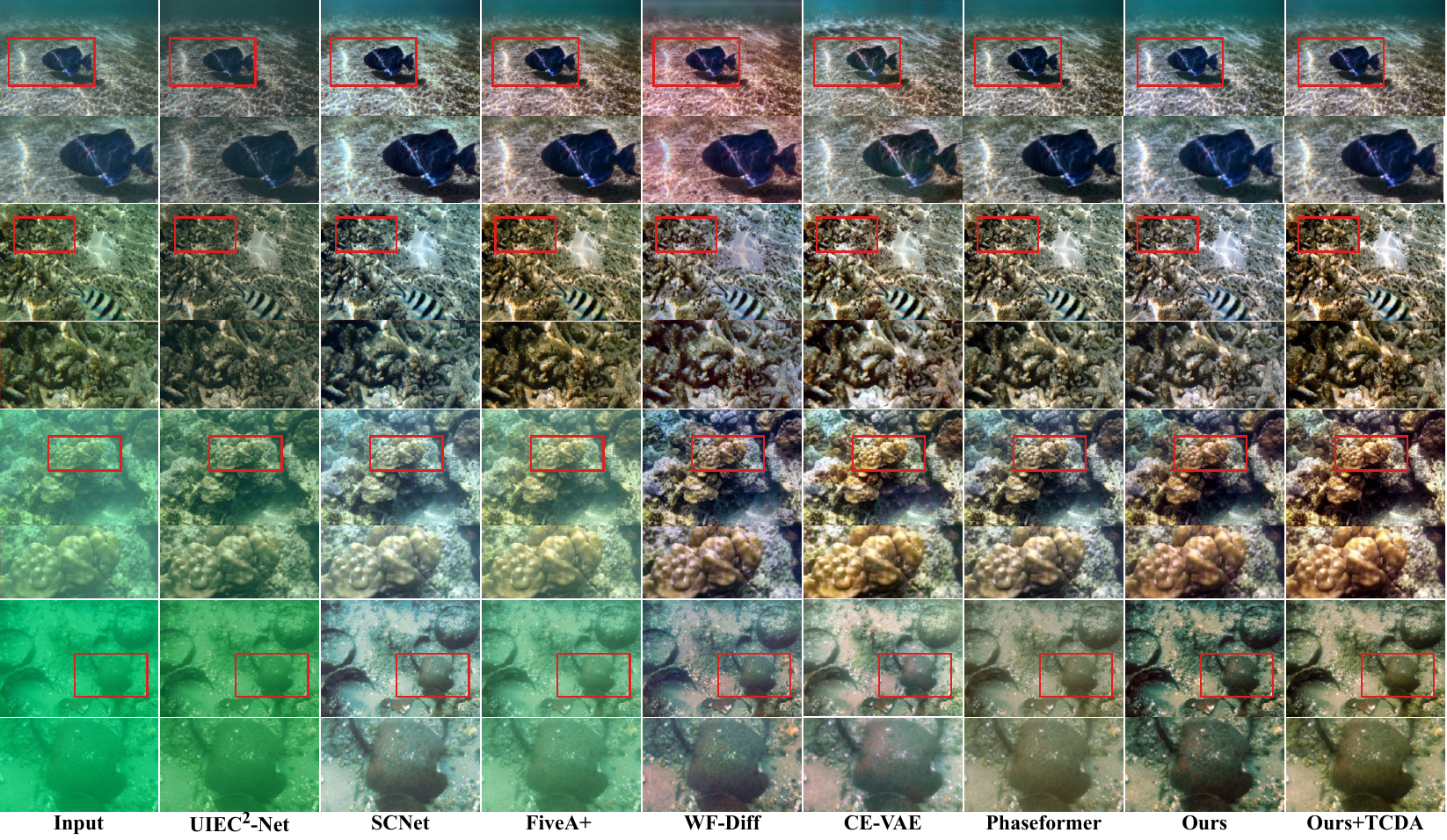}}
\caption{Visual comparisons with state-of-the-art UIE methods on the UIEB dataset. Our method consistently outperforms state-of-the-art UIE methods for underwater images with varying degrees of degradation.}
\label{Fig.method}
\end{figure*}

\subsection{Transmission-Guided One-Step Diffusion}
To bridge the gap between physics-agnostic diffusion models and restoration with greater inference efficiency than multi-step diffusion methods, we propose Transmission-Guided One-Step Diffusion (TG-Diff). Our framework builds upon the single-step SD-Turbo architecture~\cite{sauer2024adversarial}, which we adapt to the UIE task. Runtime and model-complexity comparisons are provided in Supplementary Sec.~3.1. We freeze the pre-trained weights and apply LoRA~\cite{hu2021lora} with a rank of 32 to both the U-Net and the VAE encoder.

To provide the diffusion model with a dense spatial prior on attenuation and scattering, we condition it on a transmission map $\mathcal{T}$ estimated via the UDCP method~\cite{drews2013transmission}. This strategy provides the model with a physics-inspired cue about spatially varying water properties. 
To achieve this, the map is downsampled to the latent resolution and concatenated with the image latent $z_{\text{deg}}$, forming an expanded 5-channel input $x=[z_{\text{deg}},\mathcal{T}_{\text{latent}}]$ where new channel weights are zero-initialized.

Furthermore, to preserve high-frequency details during the single-step process, we augment the VAE with multi-scale skip connections. In this process, an intermediate skip feature $S_i$ is first generated from the encoder feature $E_i$ via a $1\times1$ convolution. This feature is then added to the corresponding upsampled decoder feature $U_i$ as:
\begin{equation}
    S_i = \text{Conv}_{1\times1}(E_i), \quad U_i \leftarrow \text{Interp}(U_i) + S_i \ .
\end{equation}

\section{Experiment}
\subsection{Experimental Setup}

\noindent
\textbf{Implementation Details.}
In the TCDA pipeline, we utilize two NVIDIA A800 GPUs (80 GB) for training the degradation diffusion model. We adopt a LoRA fine-tuning strategy with a rank of 8. All the images are resized to 1024×1024 in the training process. The model is optimized using AdamW with a learning rate of $1\times10^{-4}$ and a per-GPU batch size of 1. We train the model for two epochs. During inference, we employ 28 denoising steps for text-conditioned degradation synthesis. 
For training the TG-Diff, we adopt the Stable Diffusion-Turbo~\cite{sauer2024adversarial} as our base model and fine-tune it with LoRA of rank 32. During the training process, the samples are randomly cropped into 256 $\times$ 256 patches from the original images.
We train the model for 50,000 iterations using the AdamW optimizer ($\beta_1=0.9, \beta_2=0.999$) with a constant learning rate of $1\times10^{-4}$ and a batch size of 12 per GPU. We optimize TG-Diff with the composite objective $\mathcal{L}_{\text{total}} = \mathcal{L}_{\text{L1}} + \alpha\,\mathcal{L}_{\text{LPIPS}}$, which combines the L1 reconstruction loss ($\mathcal{L}_{\text{L1}}$) and the LPIPS perceptual loss ($\mathcal{L}_{\text{LPIPS}}$), with $\alpha=3.3$ in all experiments.

\noindent
\textbf{Datasets and Evaluation Metrics.}
All images in the UIEB dataset \cite{li2019underwater} are first divided into 790 images for training and 100 images for testing. 
Based on 790 training images, we further utilize our text-conditioned degradation augmentation (TCDA) model to produce 10,270 augmented images. These augmented images are categorized into three difficulty levels (easy, moderate, and hard) with a ratio of 5:5:3.
To assess the benefit of TCDA, we train all underwater image enhancement models with and without 10,270 augmented images, and then evaluate them on the same UIEB test set.
For paired evaluations, we report PSNR~\cite{huynh2008scope}, SSIM~\cite{wang2004image}, and LPIPS~\cite{zhang2018unreasonable} to measure reconstruction fidelity. For unpaired evaluations, we employ two no-reference underwater quality metrics: UCIQE~\cite{yang2015underwater} and UIQM~\cite{panetta2015human}.
In addition, we assess performance on the reference-free UIEB C60 subset~\cite{li2019underwater} and U45 dataset~\cite{li2019fusion}, using UCIQE and UIQM to measure enhancement quality under unpaired conditions.

\subsection{Comparison with State-of-the-Arts}
We compare our approach with several state-of-the-art UIE methods, including UIEC$^{2}$-Net~\cite{wang2021uiec}, SCNet~\cite{fu2022underwater}, FiveA+~\cite{jiang2023five}, WF-Diff~\cite{zhao2024wavelet}, CE-VAE~\cite{pucci2025vae}, and Phaseformer~\cite{khan2025phaseformer}.

\noindent
\textbf{Quantitative Evaluation.}
The quantitative comparisons in Table~\ref{compare_main} show that our method achieves the strongest overall performance across three underwater benchmarks.
On the paired UIEB dataset, TG-Diff further achieves the best SSIM (0.927), PSNR (22.341), LPIPS (0.129), and UCIQE (0.613) after applying TCDA. On the reference-free C60 and U45 benchmarks, our method also obtains the highest UCIQE and UIQM scores in most cases, indicating improved performance on these reference-free metrics. TCDA improves most reported metrics across the compared methods. A discussion of occasional UIQM decreases is provided in Supplementary Sec.~3.2.

\noindent
\textbf{Qualitative Evaluation.}
Figure~\ref{Fig.method} presents visual comparisons on representative UIEB examples with different degradation levels.
Existing methods often suffer from distinct artifacts: UIEC$^2$-Net tends to leave residual haze and muted colors, SCNet and FiveA+ may introduce over-enhancement or unnatural color shifts, WF-Diff and CE-VAE sometimes fail to fully recover fine structures, and Phaseformer, although stronger, still shows less stable restoration under severe degradations.
In contrast, our TG-Diff produces clearer visibility, more natural color balance, and better local detail recovery, as highlighted in the enlarged regions.
After incorporating TCDA, “Ours+TCDA” further improves restoration fidelity, especially in heavily degraded scenes, where it removes haze more thoroughly and recovers clearer textures and structures.
These results verify that the proposed data-centric curriculum not only improves objective performance but also leads to more visually reliable enhancement results.

\begin{table}[tbp]
\centering
\setlength{\tabcolsep}{1mm}{
\caption{Ablation study on the key components of TG-Diff.} 
\label{ablation_udcp}
\begin{tabular}{@{}c@{}}
\resizebox{\columnwidth}{!}{
\begin{tabular}{cccc|ccc} 
\Xhline{1pt}
\makecell{Backbone\\(Single-step)} & \makecell{Encoder-lora\\(Trainable)} & \makecell{Skip\\Conn.} & UDCP  & $\text{SSIM}\uparrow$ & $\text{PSNR}\uparrow$ & LPIPS$\downarrow$  \\
\midrule
\checkmark &            &            &            & 0.697 & 20.374 & 0.226 \\
\checkmark & \checkmark &            &            & 0.701 & 20.516 & 0.228  \\
\checkmark &            & \checkmark &            & 0.879 & 21.266 & 0.153 \\
\checkmark &            &            & \checkmark & 0.700 & 20.436 & 0.233 \\
\checkmark & \checkmark & \checkmark &             & 0.912 & \textbf{21.767} & 0.139 \\
\checkmark & \checkmark & \checkmark & \checkmark & \textbf{0.917} & 21.723 & \textbf{0.135} \\
\Xhline{1pt}
\end{tabular}}
\end{tabular}}
\end{table}

\begin{figure}[t]
    \centering
    \includegraphics[width=\linewidth]{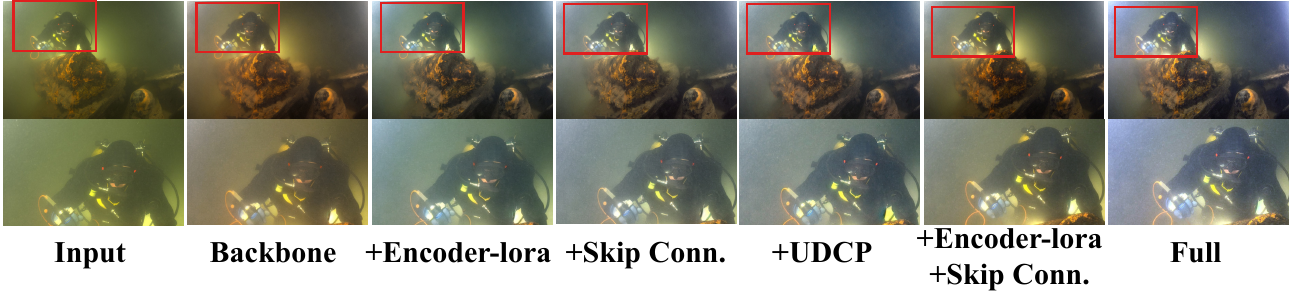}
    \caption{
    Qualitative ablation study of the proposed TG-Diff components on the UIEB dataset.
    The full model restores clearer structure, more natural illumination, and balanced color rendition.
    }
    \label{Fig.ablation-module}
\end{figure}

\begin{table}[tbp]
\centering
\setlength{\tabcolsep}{1mm}{
\caption{Quantitative comparison on LSUI~\cite{peng2023u}. Best results are in bold.} 
\label{compare_lsui}
\begin{tabular}{@{}c@{}}
\resizebox{\columnwidth}{!}{
\begin{tabular}{l|cccc} 
\Xhline{1pt}
{Methods} & $\text{SSIM}$$\uparrow$ & $\text{PSNR}$$\uparrow$ & LPIPS$\downarrow$ & UCIQE$\uparrow$  \\
\midrule
UIEC$^{2}$-Net~\cite{wang2021uiec} & 0.857 & 20.570  &  0.255  &  0.552   \\
SCNet~\cite{fu2022underwater} & 0.872 & 20.737 & 0.215 & 0.584   \\
FiveA+~\cite{jiang2023five} & 0.876 & 21.655 & 0.241 & 0.570  \\
WF-Diff~\cite{zhao2024wavelet} & 0.895 & 23.346 & 0.165  & 0.590  \\
CE-VAE~\cite{pucci2025vae} & 0.851 & 25.535 & 0.190  & 0.578   \\
Phaseformer~\cite{khan2025phaseformer} & 0.891 & 25.483 & 0.135 & 0.579   \\
\midrule
\textbf{TG-Diff (Ours)} & \textbf{0.904} & \textbf{25.734} &\textbf{0.130} &\textbf{0.592}  \\
\Xhline{1pt}
\end{tabular}}
\end{tabular}}
\end{table}

\subsection{Ablation Studies}
 
\noindent
\textbf{Effectiveness of Transmission-Guided One-Step Diffusion.}
We investigate each TG-Diff component through the quantitative and qualitative results in Table~\ref{ablation_udcp} and Figure~\ref{Fig.ablation-module}.
Starting from the single-step backbone, making the encoder trainable improves feature adaptability, while adding skip connections further enhances structural fidelity.
Incorporating the UDCP prior improves SSIM and LPIPS while  maintaining comparable PSNR, indicating complementary benefits from the transmission cue.

To further validate the effectiveness of our TG-Diff, we conduct additional comparison experiments on the LSUI dataset~\cite{peng2023u}, which contains 3,879 training images and 400 testing images.
As shown in Table~\ref{compare_lsui}, our TG-Diff consistently outperforms representative UIE networks across both reference-based and non-reference metrics, demonstrating the superiority of our network design and its strong adaptability to diverse underwater conditions.
More visual comparisons are provided in Supplementary Sec.~4.1.

\begin{figure}[t]
\centering{\includegraphics[width=0.47\textwidth]{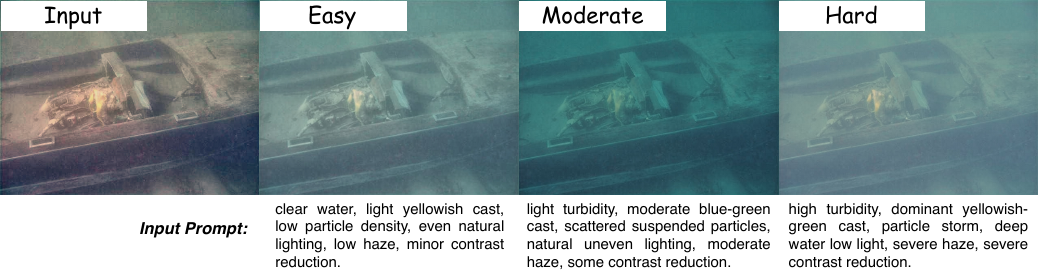}
\caption{Visualization of degradation generation across different difficulty levels. Conditioned on text prompts describing easy, moderate, and hard degradations, the model generates visually consistent degradation patterns with increasing severity.}
\label{Fig.experiment_degrade}}
\end{figure}

\begin{table}[t]
\centering
\setlength{\tabcolsep}{1.2mm}{
\caption{Ablation of degradation-difficulty sampling. Curriculum learning yields the best balanced performance.} 
\label{ablation_different_samples}
\begin{tabular}{@{}c@{}}
\resizebox{\columnwidth}{!}{
\begin{tabular}{l|cccc} 
\Xhline{1pt}
{Methods} & $\text{SSIM}$$\uparrow$ & $\text{PSNR}$$\uparrow$ & LPIPS$\downarrow$ & UCIQE$\uparrow$  \\
\midrule
TG-Diff + Easy only  & 0.901 & 21.474 & 0.149 & 0.603  \\
TG-Diff + Moderate only & 0.905 & 21.392 & 0.147 & 0.608  \\
TG-Diff + Hard only  & 0.907 & 21.404 & 0.150 & 0.603  \\
TG-Diff + Random Sample  & 0.925 & 22.248 & 0.129 & 0.608  \\
TG-Diff + Curriculum & \textbf{0.927} & \textbf{22.341} & \textbf{0.129} & \textbf{0.613}  \\
\Xhline{1pt}
\end{tabular}}
\end{tabular}}
\vspace{2mm}

\centering
\setlength{\tabcolsep}{1.2mm}{
\caption{Ablation on the augmented-to-original data ratio ($R_{\text{Aug}}$). Increasing the ratio of synthetic degradations improves overall enhancement quality.} 
\label{ablation_different_ratio}
\begin{tabular}{@{}c@{}}
\resizebox{\columnwidth}{!}{
\begin{tabular}{l|cccc} 
\Xhline{1pt}
{Methods} & $\text{SSIM}$$\uparrow$ & $\text{PSNR}$$\uparrow$ & LPIPS$\downarrow$ & UCIQE$\uparrow$  \\
\midrule
$R_{\text{Aug}}$ = 1 (only moderate) & 0.905 & 21.392 & 0.147 & 0.608 \\
$R_{\text{Aug}}$ = 3 & 0.919 & 22.126 & 0.134 & 0.611 \\
$R_{\text{Aug}}$ = 5 & 0.924 & 22.267 & 0.130 & 0.602 \\
$R_{\text{Aug}}$ = 13 & \textbf{0.927} & \textbf{22.341} & \textbf{0.129} & \textbf{0.613} \\
\Xhline{1pt}
\end{tabular}}
\end{tabular}}
\end{table}

\begin{table}[t]
\centering
\small
\setlength{\tabcolsep}{3pt}
\renewcommand{\arraystretch}{1.1}
\caption{Comparison between UWCNN-based augmentation and TCDA (moderate only).}
\label{tab:uwcnn_tcda}
\begin{tabular*}{\columnwidth}{@{\extracolsep{\fill}}l|cccc@{}}
\toprule
Method & SSIM$\uparrow$ & PSNR$\uparrow$ & LPIPS$\downarrow$ & UCIQE$\uparrow$ \\
\midrule
w/ UWCNN       & 0.855 & 19.491 & 0.185 & 0.580 \\
w/ TCDA (Ours) & 0.905 & 21.392 & 0.147 & 0.608 \\
\bottomrule
\end{tabular*}
\end{table}

\noindent
\textbf{Effectiveness of Progressive Degradation Curriculum.}
We ablate the impact of the proposed Progressive Degradation Curriculum by comparing it with training on a single difficulty level. As shown in Table~\ref{ablation_different_samples}, 
integrating curriculum learning yields the best overall performance by enabling progressive adaptation to degradations of varying difficulty.

We further investigate the effect of the augmented-to-original data ratio ($R_{\text{Aug}}$) on model performance. As shown in Table~\ref{ablation_different_ratio}, enlarging the ratio from 1× to 3× and 5× generally improves restoration fidelity   and perceptual quality.

\noindent
\textbf{Analysis of Controllable Degradation Generation.}
As shown in Figure~\ref{Fig.experiment_degrade}, our diffusion-based degradation model provides fine-grained control over degradation severity.
Conditioned on prompts with increasing difficulty, it generates consistent degradation transitions from mild color distortion to severe scattering and contrast loss.

To evaluate the distributional similarity of our synthesized degradations to real underwater images, we compare our text-driven degradation generation model with representative underwater image synthesis methods, including UWGAN~\cite{wang2019uwgan}, UWCNN~\cite{li2020underwater}, and UWNR~\cite{ye2022underwater}.
We compute the Fréchet Inception Distance (FID) between generated degraded images and real underwater samples from the UIEB test set (100 images) and the LSUI test set (400 images).
As shown in Figure~\ref{Fig.teaser}, our model achieves a lower FID score than the compared synthesis methods, suggesting closer alignment with real underwater images in the Inception feature space.

\begin{figure}
\centering{\includegraphics[width=0.5\textwidth]{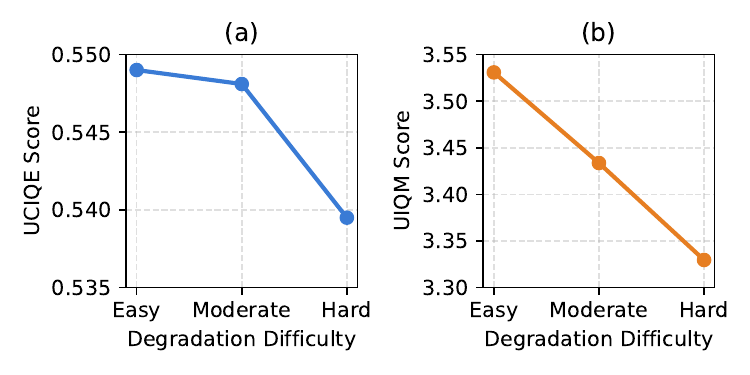}}
 \caption{Trends of non-reference underwater image quality metrics under different degradation difficulty level settings. Both UCIQE and UIQM consistently decrease as the degradation difficulty increases from easy to hard.}
\label{Fig.easy_to_hard}
\end{figure}

\noindent
\textbf{Validation of Degradation Difficulty Levels.}
We evaluate variations in the severity of our synthesized degradations using two non-reference underwater metrics: UCIQE, which reflects color fidelity and contrast, and UIQM, which measures perceptual sharpness and colorfulness.
As shown in Figure~\ref{Fig.easy_to_hard}, both scores decline consistently from easy to hard levels, indicating that the generated degradations exhibit controllable severity.

\noindent
\textbf{Effectiveness of the Pipeline.} 
As shown in Table~\ref{tab:uwcnn_tcda}, beyond FID, the TG-Diff model trained on our synthesized data achieves stronger enhancement performance than that trained on degradations generated by the representative baseline UWCNN.

\section{Conclusion}
In this work, we propose an explicit, controllable degradation modeling framework designed to alleviate the scale and diversity limitations of paired UIE data. We introduce TCDA to synthesize diverse underwater degradations with fine-grained textual control, a progressive degradation curriculum to organize these samples across attributes and difficulty levels, and TG-Diff to incorporate a transmission prior into one-step restoration. Together, these components convert controllable degradation modeling into structured supervision for robust UIE. Experiments on multiple benchmarks demonstrate improvements across most restoration and reference-free metrics, while the ablation studies examine the contributions of the curriculum, data augmentation, and transmission guidance. Future work will extend the framework to temporally consistent underwater video enhancement and broader real-world degradation conditions.

\noindent
\textbf{Data Availability Statement.} In this work, we generate augmented training data for underwater image enhancement. The data is available in: \url{https://github.com/ANTIT-Pixel/Rethinking-Underwater-Image-Enhancement}.

\noindent
\textbf{Acknowledgment.} This work is supported by Guangdong Basic and Applied Basic Research Foundation (2025A1515140021), the Nansha Key Area Science and Technology Project (No. 2024ZD006), and the Program of Guangdong Education Department.

\bibliography{main}

\vfill
\clearpage

\end{document}